\documentclass[conference]{IEEEtran}
\IEEEoverridecommandlockouts
\usepackage [latin1]{inputenc}
\usepackage{cite}
\usepackage{amsmath,amssymb,amsfonts}
\usepackage{algorithmic}
\usepackage{graphicx}
\usepackage{textcomp}
\usepackage{xcolor}
\usepackage{url}
\usepackage{comment}
\usepackage{hyperref} 
\def\BibTeX{{\rm B\kern-.05em{\sc i\kern-.025em b}\kern-.08em
    T\kern-.1667em\lower.7ex\hbox{E}\kern-.125emX}}
\begin{document}

\title{Discourse Features Enhance Detection of Document-Level Machine-Generated Content}

\author{\IEEEauthorblockN{1\textsuperscript{st} Yupei Li}
\IEEEauthorblockA{\textit{Department of Computing} \\
\textit{Imperial College London}\\
London, UK \\
yl7622@ic.ac.uk}
\and
\IEEEauthorblockN{2\textsuperscript{nd} Manual Milling}
\IEEEauthorblockA{\textit{Chair of Health Informatics} \\
\textit{Technical University of Munich}\\
Munich, Germany \\
manuel.milling@tum.de}
\and
\IEEEauthorblockN{3\textsuperscript{rd} Lucia Specia}
\IEEEauthorblockA{\textit{Department of Computing} \\
\textit{Imperial College London}\\
London, UK \\
l.specia@imperial.ac.uk}
\and
\hspace{4.5cm}
\IEEEauthorblockN{4\textsuperscript{th} Bj\"orn W. Schuller}
\IEEEauthorblockA{\hspace{5cm}\textit{Department of Computing \& Chair of Health Informatics} \\
\hspace{5cm}\textit{Imperial College London \& Technical University of Munich}\\
\hspace{5cm}London, UK \& Munich, Germany \\
\hspace{5cm}schuller@tum.de}

}

\maketitle

\begin{abstract}
The availability of high-quality APIs for Large Language Models (LLMs) has facilitated the widespread creation of Machine-Generated Content (MGC), posing challenges such as academic plagiarism and the spread of misinformation. Existing MGC detectors often focus solely on surface-level information, overlooking implicit and structural features. This makes them susceptible to deception by surface-level sentence patterns, particularly for longer texts and in texts that have been subsequently paraphrased. To overcome these challenges, we introduce novel methodologies and datasets. Besides the publicly available dataset Plagbench, we developed the paraphrased Long-Form Question and Answer (paraLFQA) and paraphrased Writing Prompts (paraWP) datasets using GPT and DIPPER, a discourse paraphrasing tool, by extending artifacts from their original versions. To better capture the structure of longer texts at document level, we propose DTransformer, a model that integrates discourse analysis through PDTB preprocessing to encode structural features. It results in substantial performance gains across both datasets -- 15.5\% absolute improvement on paraLFQA, 4\% absolute improvement on paraWP, and 1.5\% absolute improvemene on M4 compared to SOTA approaches. The data and code are available at \href{https://github.com/myxp-lyp/Discourse-Features-Enhance-Detection-of-Document-Level-Machine-Generated-Content.git}{this link}\footnote{https://github.com/myxp-lyp/Discourse-Features-Enhance-Detection-of\\
-Document-Level-Machine-Generated-Content.git}.
\end{abstract}


\section{Introduction}

\begin{figure}[htbp]
\centering
\includegraphics[width=\hsize]{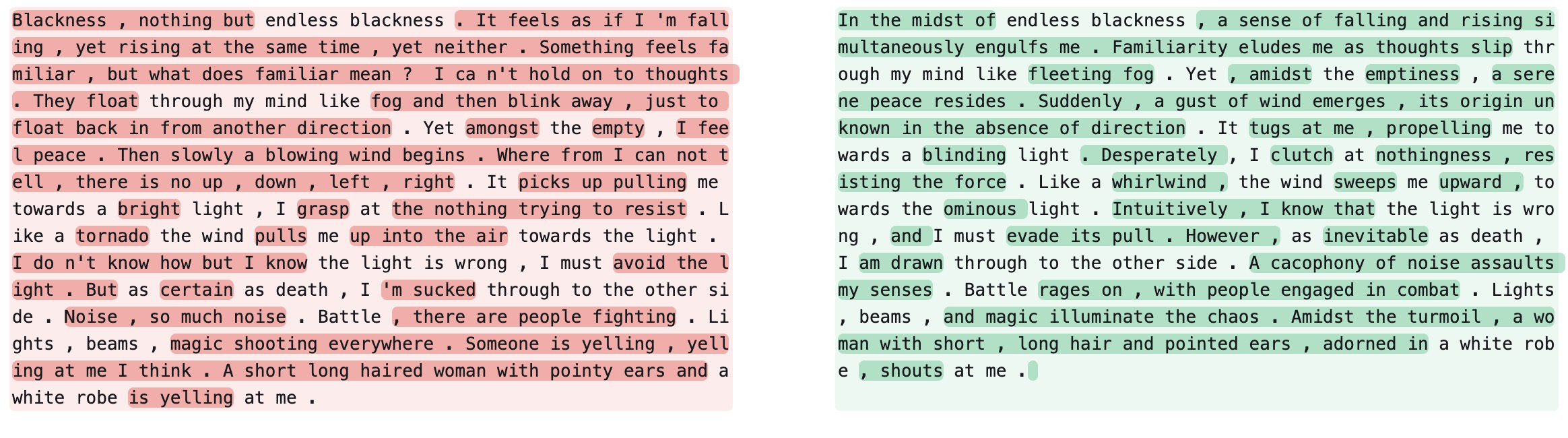}
\caption{Comparison of human-written text (left) with its corresponding GPT-revised text (right).}
\label{fig:intro case}
\end{figure}

As LLMs such as GPT-4 \cite{openai2023gpt4} and Llama3 \cite{touvron2023llama} develop rapidly and become widely utilized by the general public, Machine-Generated Content (MGC) has become increasingly abundant. In particular, people often rely on these models to revise long texts, such as essays, which results in paraphrased MGC. One example is shown in Figure \ref{fig:intro case}, where the original human-written text is highlighted in red on the left side, and the MGC in green on the right side, with differences shaded. It can be argued that content features such as fluency, readability, and clarity all improve after revision, and that the usage of LLM-usage can be hard to detect manually. 

There are well known risks associated with the misuse of MGC. \cite{crothers2023machine} conducted a survey on the applications of MGC, reporting that it may contribute to academic misconduct, the spread of fake political news, and an increase in social media spam. Our goal in this paper is the effective detection of document-level paraphrased MGC, which is rewritten by LLMs from an original human-written text. 

Automatically distinguishing MGC from Human-Produced Content (HPC) is, however, very challenging, given that MGC tends to resemble HPC, often containing at least chunks that are reproduced from human-produced training data. Currently, methods to detect  MGC involve a combination of processes, namely a Natural Language Understanding (NLU) module that extracts linguistic and contextual features, followed by a classification module which fits this to binary labels using algorithms such as Logistic Regression (LR) \cite{cox1958regression} for decision-making. 

However, these suffer from various limitations. First, current NLU models built using Transformers \cite{vaswani2017attention} such as Roberta \cite{liu2019roberta} demonstrate good performance in analysing short texts (such as sentences) but less so when processing document-level data, as shown in \cite{fu2024challenges}. They tend to struggle with retaining information if the texts are long. 
Even LLMs are limited in the length of the input texts that they can process effectively. 
{\em Document-level texts} exhibit more complex structures' relationships. Secondly, the degree of similarity between MGC and HPC is higher, especially when LLMs are used to paraphrase HPC rather than generating it from scratch. This makes distinguishing HPC and MGC challenging, as differences in linguistic content are less distinctive. Previous work focuses on identifying similarities at surface-level linguistic features, ignoring the importance of other implicit features, such as structural features, which we show can contribute to the effectiveness of a detection model. There is indeed an implicit difference in Figure \ref{fig:intro case}: the left side is more poetically structured, while the right side is a plainer narrative, though the distinction may not be obvious to the human reader.
These limitations lead to underperforming detectors in work to date. 

We propose a novel approach to automatically detect MGC which addresses the limitations described above. The aim is to better classify the texts by identifying structural features such as discourse relations. Specifically, we introduce a new model: DTransformer, which incorporates discourse features to improve detection accuracy.

As data to build and evaluate our models, we use publicly available datasets containing document-level paraphrased MGC and their HPC counterparts, 
as well as two new datasets that we created.
We show that our proposed models consistently outperform baseline and SOTA approaches, both research and commercial tools, for long-form and machine-paraphrased texts. Our contributions can be summarized as follows:

\begin{itemize}
    \item  We create paraphrased Long-Form Question and Answer (paraLFQA) and paraphrased Writing Prompts (paraWP) -- two parallel HPC-MGC corpora for model evaluation and further research in Section \ref{contribution1}
    \item  We introduce DTransformer -- a discourse-aware Transformer-based model to incorporate discourse features in detection, which leads to better performance for long-form texts in Section \ref{contribution3}.
\end{itemize}

\section{Related work}
\subsection{MGC Detection Approaches}
Given the current prominence of this topic, current work includes contributions from both academia and industry.

In the \textbf{academic} literature, two types of models are commonly used: statistical methods, also known as feature-based methods, and neural network methods. 
\cite{herbold2023ai} identified seven linguistic features and criteria for distinguishing MGC from HPC, and they concluded that MGC tends to exhibit fewer discourse and epistemic words but displays greater lexical diversity and nominalizations. \cite{choudhary2021linguistic} and Generative Language Test Room (GLTR) \cite{gehrmann2019gltr} also made decisions with linguistic features. Moreover, syntactic, sentimental, grammatical, and readability features are used to train models for fake news detection \cite{choudhary2021linguistic}. Apart from those linguistic features, there is N-gram-based methods, along with LLMs, have also been proposed for detection without requiring additional training, as demonstrated by DNA-GPT \cite{yang2023dnagptdivergentngramanalysis}.

However, feature-based methods need manual work to design the features. There have been studies comparing neural network methods and statistical methods \cite{hayawi2023imitation}, demonstrating the superior efficiency of Transformer-based models \cite{vaswani2017attention}, such as Roberta Large OpenAI Detector \cite{solaiman2019release}, the first fake news detector Grover \cite{zellers2019grover}, a zero-shot detection method DetectGpt \cite{mitchell2023detectgpt}, and watermarking-based detection \cite{fu2024watermarking}. The Roberta Large Detector has been fine-tuned with 1.5B outputs generated with GPT2. The detection is treated as a binary classification. However, it has not taken into account additional linguistic characteristics of the texts that could enhance overall performance. More importantly, 
it was trained on texts produced with GPT-2, which may retain biases from GPT-2, while current machine-generation methods have evolved (e.\,g., GPT-4). 

Additionally, Fu et al.\ \cite{fu2024watermarking} have introduced semantic-aware watermarking algorithms, while \cite{jiang2023evading} employ perturbation strategies to evade detection methods based on watermarking. Besides, Hu et al.\ \cite{hu2023radar} 
introduce another adversarial learning approach named Radar, noted for its robust transferability. Moreover, \cite{chaka2023detecting} directly utilises LLMs \cite{brown2020language} for MGC detection, but concludes that these models are currently inadequate. Conversely, \cite{chen2023gpt} combine two LLMs, BERT \cite{bert}, and T5 \cite{raffel2020exploring}, to improve detection efficacy. Moreover, recent advancements include the M4 model for multi-modality MGC detection \cite{wang2023m4} and DeBERTa \cite{he2020deberta} for monitoring mainstream news websites \cite{hanley2024machine}. 

Some efforts have been made specially on paraphrased MGC detection. \cite{wahle2022identifying} and \cite{foltynek2020detecting} proposed using pre-trained word embedding models for text representation, combined with classifiers, to predict paraphrased plagiarism in writing. Animesh et al.\  \cite{nighojkar2021improvingparaphrasedetectionadversarial} introduced an adversarial method to detect paraphrasing by leveraging the inferential properties of sentences. 
Other research has focused on developing corpus-based methods for detecting paraphrasing \cite{vrbanec2020corpus}. However, paraphrasing involves more than just substituting words and should incorporate additional detectable features. Researchers have manually crafted features, as discussed by Shahmohammadi et al.\  \cite{shahmohammadi2021paraphrase}, which align with syntactic similarity, and other factors. Unlike these methods, Li et al. \cite{li2024spottingaistouchidentifying} have developed a paraphrase detection approach for text spans. While simple, this method offers an alternative perspective.

Moreover, numerous widely utilized \textbf{commercial tools} include GPTZero\footnote{https://gptzero.me/}, Copyleaks\footnote{https://copyleaks.com/ai-content-detector}, Content at Scale\footnote{https://contentatscale.ai/ai-content-detector/}, Originality\footnote{https://originality.ai/}, Winston AI\footnote{https://gowinston.ai/}, and Hive\footnote{https://hivemoderation.com/ai-generated-content-detection}. However, these tools utilize models that are not publicly accessible, which makes understanding these tools challenging.

As discussed above, the open challenges that have not been covered in previous work are as follows: Combining implicit features with linguistic features, and detecting document-level paraphrased machine-generated texts.

\subsection{Detection of Document-level MGC}

Long texts have more complex language organization compared to shorter texts, so comprehending long texts is more challenging than shorter ones. Current transformer-based models can fail when processing longer sequences \cite{wolf2020huggingfaces} and there remains a risk of information loss due to the constrained memory and computational costs associated with self-attention operations \cite{xie2023chunkalignselectsimple}. One example is that Krishna et al. \cite{krishna2023paraphrasing} developed an attack method targeting recent detectors by employing paraphrased sentences in MGC, resulting in a significant decrease in detector accuracy. This approach utilized paraphrased sentence patterns introduced by an external model known as DIPPER \cite{krishna2023paraphrasing}, highlighting the challenges associated with detecting discourse-paraphrased sentences which combines LLMs such as T5 and a sentence shuffle strategy to generate training data for discourse paraphrasing with high semantic similarity. As a result, Krishna et al.\ discovered that discourse-paraphrased sentences are difficult to detect.

To address the aforementioned issue, \emph{hierarchical} models have been proposed that handle sentence-level information before passing on these representations for document-level processing. \cite{zhang-etal-2019-hibert} proposed the HIBERT model for document summarization. Moreover, \cite{xie2023chunkalignselectsimple} divided long sequences into different blocks, while SPECTER \cite{cohan2020specterdocumentlevelrepresentationlearning} was proposed to embed research papers by utilizing only the metadata. Longformer \cite{beltagy2020longformerlongdocumenttransformer} introduced a sparse attention mechanism that adapts to sequence length, and \cite{grail-etal-2021-globalizing} proposed to globalize sequences processed by BERT using a bidirectional GRU network to propagate information. In-context learning in OUTFOX \cite{koike2024outfoxllmgeneratedessaydetection} has been adapted for training on long essays. However, it does not explicitly learn structural organization but is instead stimulated by certain contextual relationships. Each of these models aims to address memory limitations and computational complexity by selecting representative embeddings. However, there are no dominating models utilizing discourse features.

\begin{figure}[h]
\centering
\includegraphics[width=\hsize]{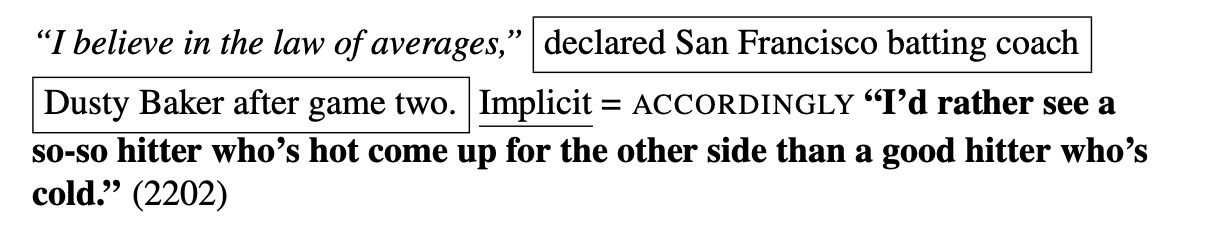}
\caption{PDTB annotation example \cite{lin2010pdtbstyled}: It shows the implicit relations, \emph{ACCORDINGLY}, between the given two sentence.}
\label{fig4}

\end{figure}

The discourse features could be represented by two annotations: Rhetorical Structure Theory Discourse Treebank (RST-DT) \cite{carlson2006rhetorical}, a dataset that captures discourse relations within text spans, represented in a tree structure and the Penn Discourse Treebank (PDTB) \cite{prasad-etal-2008-penn}, offering a straightforward representation of relations, specifying two arguments along with their relations \cite{webber2019penn} (as depicted in Figure \ref{fig4}). Annotated by human experts, PDTB's 3.0 version includes over 100 types of relations, demonstrating exceptional applications when combined with Bert models \cite{kim-etal-2020-implicit}. These annotated labels are detailed about the role of each sentence as well.
\section{Datasets}
\label{contribution1}
Currently, limited public datasets are available that focus on document-level paraphrased texts for MGC detection. For example, the recently published adversarial perturbations dataset introduced revised MGC texts with word replacements \cite{huang2024aigeneratedtextdetectorsrobust}, but the dataset does not include meaningful differences in writing style between the two versions. Its primary aim is to attack current detectors. It is difficult to adopt this dataset for our research.
Two other publicly available datasets are the Plagbench \cite{lee2024plagbenchexploringdualitylarge} and Multi-domain, and Multi-lingual Black-Box Machine-Generated Text Detection (M4) dataset \cite{wang-etal-2024-m4}. Further, in this work, we introduce two new dataset that we created based on a portion of well-known datasets used in other NLP tasks: \emph{paraLFQA} based on LFQA \cite{fan-etal-2019-eli5}, which was also used in the DIPPER dataset \cite{krishna2023paraphrasing} and \emph{paraWP} based on WritingPrompts (WP) \cite{fan2018hierarchical}.

The Plagbench dataset \cite{lee2024plagbenchexploringdualitylarge} is generated by GPT specifically for the task of plagiarism detection. We select the paraphrasing plagiarism subset, which includes human evaluation based on specially designed criteria such as semantic equivalence, adding credibility to its quality. The dataset consists of 135 data pairs. 

The M4 dataset contains a diverse array of human-written sources, including document-level texts from WikiHow, and others. It also includes MGC versions produced by a wide range of generators, such as BLOOMz \cite{bloomz2022} and ChatGPT. These machine-generated texts often differ in meaning from the original sources because the prompts used are not direct paraphrases but involve expansions, such as the prompt \emph{``Write a peer review based on the article."} It has been extensively utilized as a benchmark in MGC detection research. The dataset has a total of 133,551 text samples. 

For the paraLFQA dataset, the original answers from LFQA are considered as HPC. MGC was produced using the DIPPER model from the HPCs, resulting in discourse-paraphrased-style sentences, which we present as a new dataset. In accordance with the methodology outlined in the original paper, we apply T5 as the tokenizer. We selected a lexicon code of 40 and a re-order code of 100. These values represent that 40\% of the lexicon and 100\% of the order have been modified. These hyperparameters were chosen to emphasize changes in the organization of the language within the dataset. The dataset has around 2,000 MGC-HPC paragraph pairs with around 250-350 words in each paragraph. This serves as a control dataset where the semantic features are similar between the paraphrased and original content, but the structure organisation is different. One example is shown in Table \ref{tab:case}. It demonstrates that the sentence structures are entirely different, while the overall meaning remains the same.
\begin{table*}[htbp]
  \centering
  
  \begin{tabular}{p{7.5cm}p{7.5cm}}
    \hline
    Original Case (HPC) & Paraphrased Case (HPC)  \\
    \hline
    Pushing the bulbous part does NOT create a vacuum, no. It just pushed the air in the balloon towards the end of the pipette. If you release your fingers and the balloon wants to return to it's initial round form (we come to that later on again), then the fluid at the entrance gets pushed in. The atmosphere that is in contact with the fluid pushes the fluid into the pipette. It does this because the balloon has some tension in its surface, and it wants to be round and flat, and not squeezed together. So as it opens up, the additional volume created within the balloon is filled due to the pressure from the outside atmosphere pushing onto the surface of the fluid, which in turn gets pushed into the pipette Awesome question, BTW! & First of all, you're not creating a vacuum by pushing on the bulb. What you're doing is pushing the air in the bulb towards the tip of the tube. Now, when you let go of the bulb and it wants to go back to its original shape (we'll come back to that), the liquid in the tip will be pushed in. Basically, the liquid in the bulb has a certain tension to it and, being a bulb, wants to be flattened, not squeezed, so, when it's flattened, it fills with air from outside and the liquid is pushed into the tube. A very good question.  \\
    \hline
  \end{tabular}
  \vspace{0.1cm}
  \caption{One case for comparison between original (HPC) and discourse paraphrased (MGC) text in paraLFQA.}
  \label{tab:case}
\end{table*}

For the paraWP dataset, unlike paraLFQA focusing especially on discourse paraphrase, paraWP perform a generally much longer document-level dataset than paraLFQA. Specifically, human-written stories from the provided prompts are considered as HPC. The MGC paraphrases are produced by GPT-3.5-turbo with the following prompt: ``\emph{Summarize the text. Expand the idea into new text. Keep the original narrative, such as he, she, I, or their names. Return the same number of words as the input texts.}" In this way, the dataset has one-to-one matched text pairs with HPC and MGC, which are regarded as paraphrased-style versions. We have created 166,247 documents both for HPC and MGC in the training set, and 12,800 and 12,316 documents for the validation and test sets, respectively. We used the BPE tokenizer to process the data. The vocabulary size of the BPE modelling is 8,016. We have also investigated linguistic features of the dataset in Appendix \ref{apeendix:a}.

We also selected a portion of the paraWP dataset for revision to introduce surface-level word twist attacks by simply replacing certain words. This has been shown to be challenging for other detectors \cite{sadasivan2024aigeneratedtextreliablydetected}. We manually performed the revision task to generate a small number of modified paraWP cases by: (a) incorporating present or past participles as adverbial clauses and using appositives, and (b) employing metaphors and parallel structures in sentences. This approach systematically reduces the frequency of adverbial clauses with -ing and -ed forms, appositives, and parallel sentence structures, which are surface-level word twist attacks.

Both paraLFQA and paraWP are downloadable\footnote{https://drive.google.com/file/d/1fvsWwHKplf0-n6PnwbxIRmR6jgu62nRi/view?usp=sharing} and are licensed under a Creative Commons Attribution 4.0 International License.

\section{Model: DTransformer}
\label{contribution3}
\begin{figure}[h]
\centering
\includegraphics[width=\linewidth]{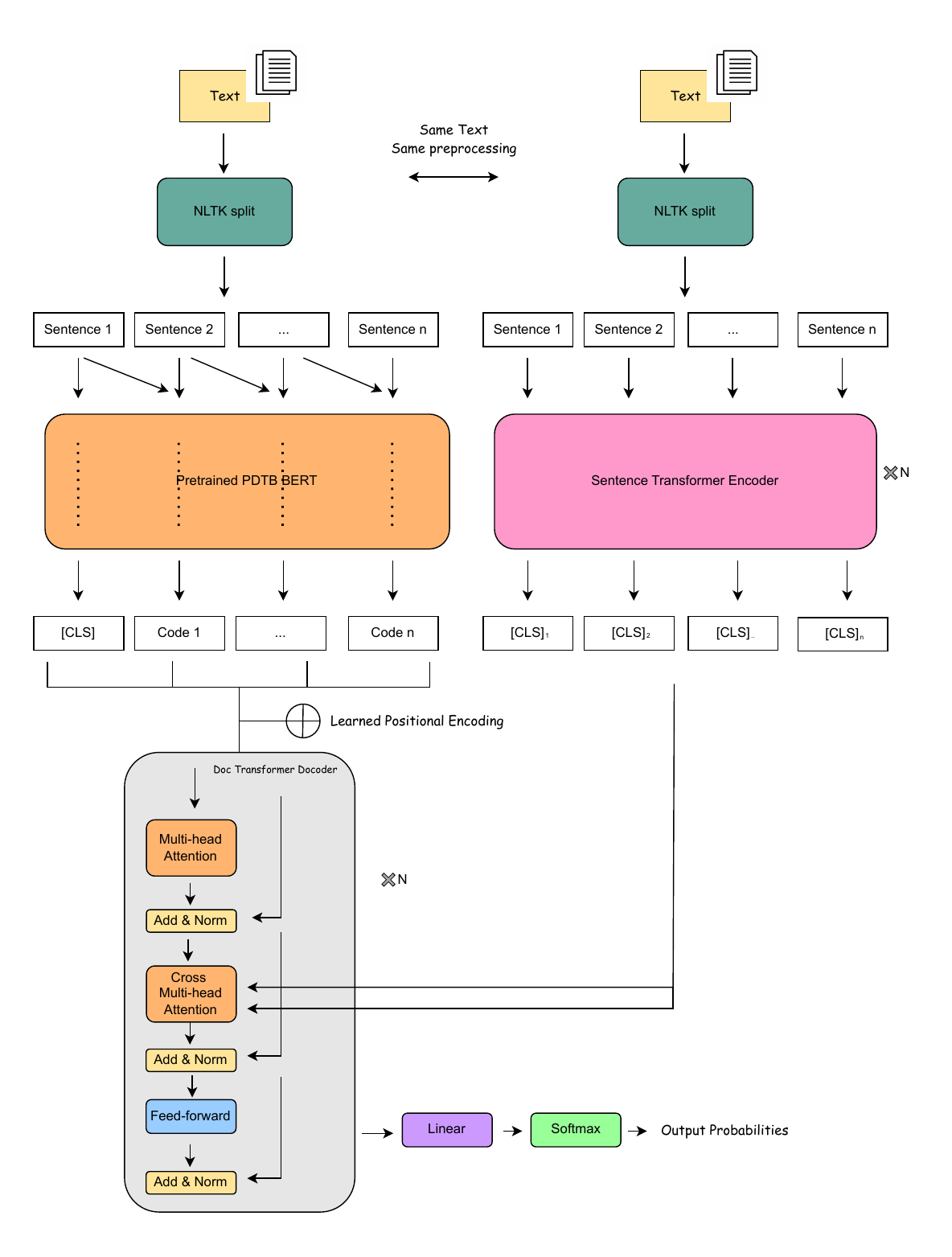}
\caption{DTransformer model: it incorporates both structural and semantic features. It first splits the paragraphs and employs a hierarchical model to capture document semantic features. Then, using a cross-attention mechanism, it analyses discourse features for improved classification.}
\label{fig6}
\end{figure}

This section introduces our method, specifically designed to address the second challenge: detecting document-level MGC texts paraphrased from HPC. The DTransformer model focuses on integrating structural features with semantic features, aiming to detect MGC by identifying the ways humans and machines organise text blocks, such as paragraphs. This model is inspired by the discourse information extraction method with PDTB annotation and the effects of structural features discussed in the Multi-Method Qualitative Text Framework \cite{alejandro2023multi}. The model is depicted in Figure \ref{fig6}, with detailed sub-modules discussed in the following paragraphs.

We use the NLTK library to split the document into individual sentences. 
Each sentence is then processed through a traditional $N$-layer Transformer Encoder. This results in the start token ($[CLS]$) representing the entire sentence's information, which encapsulates the semantic features. 

As the encoder extracts the content features, the same sentences will be processed using a BERT model pre-trained on PDTB-annotated texts to identify discourse features. This process involves taking two consecutive sentences as input and producing a single discourse relation as output. The first sentence will generate one $[CLS]$ token, and the following sentence will correspond to one discourse relation, referred to as a `discourse code', representing the relationship between itself and its preceding sentence. This module will provide $n$ discourse codes to represent the structure of the document. 

After preparing the structural and content features, the code sequences will be input into the decoder. The decoder is similar to a standard Transformer decoder block, with modifications made to the cross-attention mechanism. We use $[CLS]$ to generate $Q$, $Code$ to generate $K,V$. The final layer of the decoder will output a probability using a linear network layer and the softmax function to determine the final classification as either MGC or HPC.

\section{Implementation Details}

We prepared the M4, paraLFQA, and paraWP datasets by randomly splitting each into 80\,\% for training, 10\,\% for validation, and 10\% for testing with a fixed random seed. The Plagbench dataset was randomly divided into 60\,\% for training, 20\,\% for validation, and 20\,\% for testing due to the small size of the Plagbench dataset, the standard 10\,\% test split makes it difficult to accurately measure the model's performance. Since the MGC is generated from HPC, we ensure that the MGC and HPC are split equally to achieve a balanced dataset.

We built the RoBERTa-large baseline using the pre-trained model from Hugging Face\footnote{https://huggingface.co/openai-community/roberta-large-openai-detector}. We adjusted the maximum sentence length from 512 to 1024 to accommodate long-form texts. The model was fine-tuned with a batch size of 2 and a learning rate of 2e-5.


For DTransformer, we initialized our model parameters using the bart-large-cnn model\footnote{https://github.com/inferless/Facebook-bart-cnn} and Kaiming uniform initialization \cite{he2015delving}, respectively. The hyperparameters and optimizer selection are listed in Table \ref{tab: DTransformer hyperparameters}. 
\begin{table}[htbp]
  \centering
  \begin{tabular}{cc}
    \hline
    Hyperparameter & Value  \\
    \hline
    embedding dim & 512 \\
    feedforward embedding dim & 2048 \\
    encoder layer & 6 \\
    decoder layer & 6 \\
    encoder attention heads & 8 \\
    decoder attention heads & 8 \\
    dropout & 0.1 \\
    activation function & sigmoid \\
    max-sentence-length  & 128 \\
    max-paragraph-length  & 128 \\
    batch & 4/16\\
    learning rate & 2e-6\\
    epochs  & 20\\
    early stop epochs  & 5\\
    \hline
    optimizer   & Adam\\
    \hline
    tokenizer   & white-space\\
    \hline
  \end{tabular}
  \caption{Hyperparameters, tokenizer and optimizer for DTransformer.}
  \label{tab: DTransformer hyperparameters}
\end{table}
The batch sizes differ between the two datasets of DTransformer training is due to GPU memory limitations. We use a batch size of 4 for paraWP and 16 for paraLFQA, reflecting the varying dataset sizes, with paraLFQA being slightly shorter in length. The device we use is a Tesla V100-PCIE-32GB.  

\section{Evaluation}

We present the classification accuracy for both our models, alongside the results from the baseline and SOTA academic and commercial approaches, namely RoBERTa Large, GPTZero, Copyleaks, and Winston AI. Given that the dataset is balanced (50\% MCG and 50\% HPC), accuracy is sufficient as an evaluation metric (a majority class baseline would achieve 50\% accuracy). The results are shown in Table \ref{tab:results}.
The RoBERTa Large detector classified most cases as HPC, resulting in an accuracy around 50\%. Subsequently, we discuss the results and analyze the effectiveness of our proposed models.
\begin{table*}[t]
    \centering
    \small
    \caption{Binary classification accuracy for DTransformer compared with baseline and SOTA commercial models on both datasets.}
    \label{tab:results}
    \begin{tabular}{lccccc}
        \hline
        \textbf{Model} & \textbf{paraLFQA} & \textbf{paraWP} & \textbf{Revised paraWP} & \textbf{Plagbench} & \textbf{M4-wikihow} \\
        \hline
        \textbf{GPTZero} & 0.52 & 0.95 & 0 & 0.69 & 0.57\\
        \textbf{Copyleaks} & 0.50 & 0.67 & 0.4 & 0.59 & 0.91 \\
        \textbf{Winston ai} & 0.51 & 0.50 & 0.6 & 0.43 & 0.93 \\ \hline
        \textbf{RoBERTa Large} & 0.50 & 0.50 & 0.2 & 0.54 & 0.49 \\
        
        \textbf{DTransformer (PDTB2.0)} & 0.67 & \textbf{0.99} & \textbf{0.7} & \textbf{0.69} & \textbf{0.95}\\
        \textbf{DTransformer (PDTB3.0)} & \textbf{0.70} & \textbf{0.99} & \textbf{0.7} & 0.61 & 0.94 \\
        \hline
    \end{tabular}
\end{table*}

The results presented in Table \ref{tab:results} demonstrate that the DTransformer outperforms all other models across datasets, especially the paraLFQA, showing its efficacy in extracting features from document-level texts. For ParaLFQA, the discourse paraphrasing is particularly high, such that DTransformer could take advantage of discourse information while other models fail to pay attention to this aspect. These performances highlight the significance of incorporating discourse features. The criticality of discourse features is further validated through comparative experimentation using PDTB 2.0 and 3.0, where PDTB 3.0 consistently outperforms 2.0. This superiority is attributable to PDTB 3.0's more comprehensive discourse annotations, which better represent discourse features effectively. Specifically, the best-performing DTransformer model utilizing PDTB 3.0 shows a 4\,\%, 15.5\,\% 1.5\,\% absolute improvement over existing SOTA models on the paraWP, paraLFQA, and M4 datasets, respectively, which is strong evidence for DTransformer's ability for general document-level detection.

Our results reveal that DTransformer exhibits high performance on the paraWP dataset, achieving an accuracy of 99\,\%. This indicates that DTransformer possesses a remarkable capability for detecting MGC, even within the constraints of a challenging dataset. Furthermore, the design of the generation prompt plays a crucial role. Specifically, we ensured that the generated MGCs were predominantly distinct from their corresponding HPCs at the discourse feature level, while controlling for other variables such as semantic meaning and word count. This alignment with DTransformer's capabilities accounts for the exceptionally high accuracy observed.

We also performed a case study, shown in Table \ref{tab: DTransformer case study} to assess the effectiveness of the PDTB. For this evaluation, we employed the DTransformer with PDTB 3.0 as the representative model. The case study analyzed a paragraph from the paraLFQA dataset, originally labeled as MGC. This paragraph was misclassified by GPTZero as HPC, but it was correctly classified as HPC by DTransformer.
\begin{table*}[htbp]
    \centering
    \small
    \caption{DTransformer test example: A MGC-paragraph from paraLFQA is selected and its corresponding PDTB code prediction is presented. DTransformer made correct classifications while GPTZero classified falsely.}
    \label{tab: DTransformer case study}
    \begin{tabular}{p{7.5cm}p{7.5cm}}
        \hline
        \textbf{Text} &  \textbf{PDTB code} \\
        \hline
        Basically, they are related to the amount of ``advantage" which a lever and a winch give you. & [CLS]\\
        A lever or a pulley can make a heavy object move, because they transform a weak but large movement into a small but strong movement. & Expansion.Level-of-detail \\
        For example, by operating a come-along, you move a meter of the long lever with your arm, which is transformed into moving the heavy load one centimeter. & Expansion.Instantiation \\
        If you do this (divide the distance by a hundred) you multiply the force by a hundred, and so you can handle a weight you could move with 50 kilograms by hand, but with the help of the come-along you can handle a weight of 5,000 kilograms. & Expansion.Level-of-detail \\
        When using a ratchet strap to secure a load, the lever you move is much shorter and therefore moves less. & Expansion.Level-of-detail \\
        But you have to know that if the end of the lever moves ten centimeters and the ratchet strap moves one centimeter, your force is multiplied by ten. & Contingency.Cause\\
        So you convert your strength of 50 kilograms into a strength that can lift 500 kilograms. & Contingency.Cause \\
        When the strap is pulled to the point where it feels as if 500 kilograms are pulling on it, your strength of 50 kilograms is no longer enough to tighten it any more. & Expansion.Level-of-detail \\
        So, in essence, a come-along has a longer lever to give you a larger advantage. & Contingency.Cause
        \\
        \hline
    \end{tabular}
\end{table*}

The PDTB codes are intuitively accurate. For instance, the `expansion.level-of-detail' code corresponds to the elaboration on the `lever' in the second sentence, following the `lever' discussed in the previous sentence. Additionally, the `contingency.cause' code is marked by sentences containing conjunctions such as `but' and `so' to indicate cause and effect relationships. The `expansion.instantiation' code is identified by the presence of the phrase `for example.' These examples demonstrate that PDTB coding aligns well with human judgment, providing evidence for the effectiveness of discourse features in detecting MGC. Conversely, we hypothesize that GPTZero produced incorrect classifications due to its relative emphasis on writing style and other semantic features rather than on discourse features.

\vspace{-0.2cm}
In addition, we conducted an \textbf{ablation study} focusing on discourse features. To successfully extract document-level semantic meaning as DTransformer does, we trained the original HiBERT model \cite{zhang-etal-2019-hibert} on the paraWP dataset. However, the predicted accuracy with the hierarchical architecture model alone is 0.53 and 0.49 for the PDTB3.0 and PDTB2.0 annotated datasets, respectively on paraWP. This finding highlights the importance of incorporating both discourse and semantic features for effective document-level MGC detection. Additionally, it highlights that the accuracy gain for DTransformer is primarily attributable to discourse features.

In light of the models' high performance, we aim to evaluate the \textbf{scalability} of our models. We conducted five independent experiments for each model, utilizing different random seeds. DTransformer with PDTB 2.0 resulted in a mean test accuracy of 0.67 and a standard deviation of 0.03 on the paraLFQA dataset. The DTransformer with PDTB 3.0 produced 
a mean test accuracy of 0.70 and a standard deviation of 0.02. These results indicate the high mean accuracy and consistency of both models, demonstrating good scalability. The lower standard deviation of PDTB 3.0 suggests more stable performance across different runs. Additionally, the high accuracy of the revised paraWP serves as evidence that DTransformer is highly robust against surface-level attacks.

Based on this analysis, we argue that our findings are theoretically supported: machines struggle to maintain a stable discourse structure in document-level text generation, whereas humans can preserve a clear central theme when writing long texts. This is consistent with previous research suggesting that LLMs cannot effectively process and comprehend long documents. Instead, they tend to segment texts into chunks and process them sequentially, leading to a loss of contextual coherence \cite{agrawal2024can, abedu2024llm}. Our approach is effective because we explicitly provide this structural information to the model.
\section{Conclusions} 

In this work, we introduced two new datasets aiming to provide insights into the challenges in current work aimed at detecting MGC, particularly for document-level paraphrased machine-generated texts: paraLFQA, emphasizing the discourse difference, and paraWP, a long dataset with high similarity between HPC and MGC. Further, we introduced a novel architecture, DTransformer, which utilises discourse features, outperforming current commercial and non-commercial models across the datasets paraLFQA, paraWP, Plagbench, and M4. Future work could focus on training on more datasets for out of domain tests, enhancing robustness of the model.



\bibliographystyle{IEEEtran}
\bibliography{reference}

\begin{thebibliography}{10}
\providecommand{\url}[1]{#1}
\csname url@samestyle\endcsname
\providecommand{\newblock}{\relax}
\providecommand{\bibinfo}[2]{#2}
\providecommand{\BIBentrySTDinterwordspacing}{\spaceskip=0pt\relax}
\providecommand{\BIBentryALTinterwordstretchfactor}{4}
\providecommand{\BIBentryALTinterwordspacing}{\spaceskip=\fontdimen2\font plus
\BIBentryALTinterwordstretchfactor\fontdimen3\font minus \fontdimen4\font\relax}
\providecommand{\BIBforeignlanguage}[2]{{%
\expandafter\ifx\csname l@#1\endcsname\relax
\typeout{** WARNING: IEEEtran.bst: No hyphenation pattern has been}%
\typeout{** loaded for the language `#1'. Using the pattern for}%
\typeout{** the default language instead.}%
\else
\language=\csname l@#1\endcsname
\fi
#2}}
\providecommand{\BIBdecl}{\relax}
\BIBdecl

\bibitem{openai2023gpt4}
OpenAI, ``Gpt-4 technical report,'' 2023.

\bibitem{touvron2023llama}
H.~Touvron, T.~Lavril, G.~Izacard, X.~Martinet, M.-A. Lachaux, T.~Lacroix, B.~RoziÃ¨re, N.~Goyal, E.~Hambro, F.~Azhar, A.~Rodriguez, A.~Joulin, E.~Grave, and G.~Lample, ``Llama: Open and efficient foundation language models,'' 2023.

\bibitem{crothers2023machine}
E.~Crothers, N.~Japkowicz, and H.~L. Viktor, ``Machine-generated text: A comprehensive survey of threat models and detection methods,'' \emph{IEEE Access}, 2023.

\bibitem{cox1958regression}
D.~R. Cox, ``The regression analysis of binary sequences,'' \emph{Journal of the Royal Statistical Society: Series B (Methodological)}, vol.~20, no.~2, pp. 215--232, 1958.

\bibitem{vaswani2017attention}
A.~Vaswani, N.~Shazeer, N.~Parmar, J.~Uszkoreit, L.~Jones, A.~N. Gomez, Å.~Kaiser, and I.~Polosukhin, ``Attention is all you need,'' in \emph{Advances in Neural Information Processing Systems (NeurIPS)}, 2017.

\bibitem{liu2019roberta}
Y.~Liu, M.~Ott, N.~Goyal, J.~Du, M.~Joshi, D.~Chen, O.~Levy, M.~Lewis, L.~Zettlemoyer, and V.~Stoyanov, ``Roberta: A robustly optimized bert pretraining approach,'' in \emph{Proceedings of the 2019 Conference on Empirical Methods in Natural Language Processing and the 9th International Joint Conference on Natural Language Processing (EMNLP-IJCNLP)}, 2019, pp.~--.

\bibitem{fu2024challenges}
Y.~Fu, ``Challenges in deploying long-context transformers: A theoretical peak performance analysis,'' 2024.

\bibitem{herbold2023ai}
S.~Herbold, A.~Hautli-Janisz, U.~Heuer, Z.~Kikteva, and A.~Trautsch, ``Ai, write an essay for me: A large-scale comparison of human-written versus chatgpt-generated essays,'' 2023.

\bibitem{choudhary2021linguistic}
A.~Choudhary and A.~Arora, ``Linguistic feature based learning model for fake news detection and classification,'' \emph{Expert Systems with Applications}, vol. 169, p. 114171, 2021.

\bibitem{gehrmann2019gltr}
S.~Gehrmann, H.~Strobelt, and A.~M. Rush, ``Gltr: Statistical detection and visualization of generated text,'' \emph{arXiv preprint arXiv:1906.04043}, 2019.

\bibitem{yang2023dnagptdivergentngramanalysis}
\BIBentryALTinterwordspacing
X.~Yang, W.~Cheng, Y.~Wu, L.~Petzold, W.~Y. Wang, and H.~Chen, ``Dna-gpt: Divergent n-gram analysis for training-free detection of gpt-generated text,'' 2023. [Online]. Available: \url{https://arxiv.org/abs/2305.17359}
\BIBentrySTDinterwordspacing

\bibitem{hayawi2023imitation}
K.~Hayawi, S.~Shahriar, and S.~S. Mathew, ``The imitation game: Detecting human and ai-generated texts in the era of large language models,'' \emph{arXiv preprint arXiv:2307.12166}, 2023.

\bibitem{solaiman2019release}
I.~Solaiman, M.~Brundage, J.~Clark, A.~Askell, A.~Herbert-Voss, J.~Wu, A.~Radford, G.~Krueger, J.~W. Kim, S.~Kreps, M.~McCain, A.~Newhouse, J.~Blazakis, K.~McGuffie, and J.~Wang, ``Release strategies and the social impacts of language models,'' 2019.

\bibitem{zellers2019grover}
R.~Zellers, A.~Holtzman, H.~Rashkin, Y.~Bisk, A.~Farhadi, F.~Roesner, and Y.~Choi, ``Grover: A state-of-the-art defense against neural fake news,'' in \emph{Proceedings of the 28th International Joint Conference on Artificial Intelligence (IJCAI)}, 2019.

\bibitem{mitchell2023detectgpt}
E.~Mitchell, Y.~Lee, A.~Khazatsky, C.~D. Manning, and C.~Finn, ``Detectgpt: Zero-shot machine-generated text detection using probability curvature,'' \emph{arXiv preprint arXiv:2301.11305}, 2023.

\bibitem{fu2024watermarking}
Y.~Fu, D.~Xiong, and Y.~Dong, ``Watermarking conditional text generation for ai detection: Unveiling challenges and a semantic-aware watermark remedy,'' in \emph{Proceedings of the AAAI Conference on Artificial Intelligence}, vol.~38, no.~16, 2024, pp. 18\,003--18\,011.

\bibitem{jiang2023evading}
Z.~Jiang, J.~Zhang, and N.~Z. Gong, ``Evading watermark based detection of ai-generated content,'' in \emph{Proceedings of the 2023 ACM SIGSAC Conference on Computer and Communications Security}, 2023, pp. 1168--1181.

\bibitem{hu2023radar}
X.~Hu, P.-Y. Chen, and T.-Y. Ho, ``Radar: Robust ai-text detection via adversarial learning,'' \emph{Advances in Neural Information Processing Systems}, vol.~36, pp. 15\,077--15\,095, 2023.

\bibitem{chaka2023detecting}
C.~Chaka, ``Detecting ai content in responses generated by chatgpt, youchat, and chatsonic: The case of five ai content detection tools,'' \emph{Journal of Applied Learning and Teaching}, vol.~6, no.~2, 2023.

\bibitem{brown2020language}
\BIBentryALTinterwordspacing
T.~Brown, B.~Mann, N.~Ryder, M.~Subbiah, J.~Kaplan, P.~Dhariwal, A.~Neelakantan, P.~Shyam, G.~Sastry, A.~Askell \emph{et~al.}, ``Language models are few-shot learners,'' \emph{arXiv preprint arXiv:2005.14165}, 2020. [Online]. Available: \url{https://arxiv.org/abs/2005.14165}
\BIBentrySTDinterwordspacing

\bibitem{chen2023gpt}
Y.~Chen, H.~Kang, V.~Zhai, L.~Li, R.~Singh, and B.~Raj, ``Gpt-sentinel: Distinguishing human and chatgpt generated content,'' \emph{arXiv preprint arXiv:2305.07969}, 2023.

\bibitem{bert}
J.~Devlin, M.-W. Chang, K.~Lee, and K.~Toutanova, ``Bert: Pre-training of deep bidirectional transformers for language understanding,'' in \emph{Conference of the North American Chapter of the Association for Computational Linguistics}, 2019.

\bibitem{raffel2020exploring}
\BIBentryALTinterwordspacing
C.~Raffel, N.~Shazeer, A.~Roberts, K.~Lee, S.~Narang, M.~Matena, Y.~Zhou, W.~Li, and P.~J. Liu, ``Exploring the limits of transfer learning with a unified text-to-text transformer,'' \emph{Journal of Machine Learning Research}, vol.~21, no. 140, pp. 1--67, 2020. [Online]. Available: \url{http://jmlr.org/papers/v21/20-074.html}
\BIBentrySTDinterwordspacing

\bibitem{wang2023m4}
Y.~Wang, J.~Mansurov, P.~Ivanov, J.~Su, A.~Shelmanov, A.~Tsvigun, C.~Whitehouse, O.~M. Afzal, T.~Mahmoud, T.~Sasaki \emph{et~al.}, ``M4: Multi-generator, multi-domain, and multi-lingual black-box machine-generated text detection,'' \emph{arXiv preprint arXiv:2305.14902}, 2023.

\bibitem{he2020deberta}
\BIBentryALTinterwordspacing
P.~He, X.~Liu, J.~Gao, and W.~Chen, ``Deberta: Decoding-enhanced bert with disentangled attention,'' \emph{arXiv preprint arXiv:2006.03654}, 2020. [Online]. Available: \url{https://arxiv.org/abs/2006.03654}
\BIBentrySTDinterwordspacing

\bibitem{hanley2024machine}
H.~W. Hanley and Z.~Durumeric, ``Machine-made media: Monitoring the mobilization of machine-generated articles on misinformation and mainstream news websites,'' in \emph{Proceedings of the International AAAI Conference on Web and Social Media}, vol.~18, 2024, pp. 542--556.

\bibitem{wahle2022identifying}
J.~P. Wahle, T.~Ruas, T.~Folt{\`y}nek, N.~Meuschke, and B.~Gipp, ``Identifying machine-paraphrased plagiarism,'' in \emph{International Conference on Information}.\hskip 1em plus 0.5em minus 0.4em\relax Springer, 2022, pp. 393--413.

\bibitem{foltynek2020detecting}
T.~Folt{\`y}nek, T.~Ruas, P.~Scharpf, N.~Meuschke, M.~Schubotz, W.~Grosky, and B.~Gipp, ``Detecting machine-obfuscated plagiarism,'' in \emph{International conference on information}.\hskip 1em plus 0.5em minus 0.4em\relax Springer, 2020, pp. 816--827.

\bibitem{nighojkar2021improvingparaphrasedetectionadversarial}
\BIBentryALTinterwordspacing
A.~Nighojkar and J.~Licato, ``Improving paraphrase detection with the adversarial paraphrasing task,'' 2021. [Online]. Available: \url{https://arxiv.org/abs/2106.07691}
\BIBentrySTDinterwordspacing

\bibitem{vrbanec2020corpus}
T.~Vrbanec and A.~Me{\v{s}}trovi{\'c}, ``Corpus-based paraphrase detection experiments and review,'' \emph{Information}, vol.~11, no.~5, p. 241, 2020.

\bibitem{shahmohammadi2021paraphrase}
H.~Shahmohammadi, M.~Dezfoulian, and M.~Mansoorizadeh, ``Paraphrase detection using lstm networks and handcrafted features,'' \emph{Multimedia Tools and Applications}, vol.~80, no.~4, pp. 6479--6492, 2021.

\bibitem{li2024spottingaistouchidentifying}
\BIBentryALTinterwordspacing
Y.~Li, Z.~Wang, L.~Cui, W.~Bi, S.~Shi, and Y.~Zhang, ``Spotting ai's touch: Identifying llm-paraphrased spans in text,'' 2024. [Online]. Available: \url{https://arxiv.org/abs/2405.12689}
\BIBentrySTDinterwordspacing

\bibitem{wolf2020huggingfaces}
T.~Wolf, L.~Debut, V.~Sanh, J.~Chaumond, C.~Delangue, A.~Moi, P.~Cistac, T.~Rault, R.~Louf, M.~Funtowicz, J.~Davison, S.~Shleifer, P.~von Platen, C.~Ma, Y.~Jernite, J.~Plu, C.~Xu, T.~L. Scao, S.~Gugger, M.~Drame, Q.~Lhoest, and A.~M. Rush, ``Huggingface's transformers: State-of-the-art natural language processing,'' 2020.

\bibitem{xie2023chunkalignselectsimple}
\BIBentryALTinterwordspacing
J.~Xie, P.~Cheng, X.~Liang, Y.~Dai, and N.~Du, ``Chunk, align, select: A simple long-sequence processing method for transformers,'' 2023. [Online]. Available: \url{https://arxiv.org/abs/2308.13191}
\BIBentrySTDinterwordspacing

\bibitem{krishna2023paraphrasing}
K.~Krishna, Y.~Song, M.~Karpinska, J.~Wieting, and M.~Iyyer, ``Paraphrasing evades detectors of ai-generated text, but retrieval is an effective defense,'' 2023.

\bibitem{zhang-etal-2019-hibert}
\BIBentryALTinterwordspacing
X.~Zhang, F.~Wei, and M.~Zhou, ``{HIBERT}: Document level pre-training of hierarchical bidirectional transformers for document summarization,'' in \emph{Proceedings of the 57th Annual Meeting of the Association for Computational Linguistics}, A.~Korhonen, D.~Traum, and L.~M{\`a}rquez, Eds.\hskip 1em plus 0.5em minus 0.4em\relax Florence, Italy: Association for Computational Linguistics, Jul. 2019, pp. 5059--5069. [Online]. Available: \url{https://aclanthology.org/P19-1499}
\BIBentrySTDinterwordspacing

\bibitem{cohan2020specterdocumentlevelrepresentationlearning}
\BIBentryALTinterwordspacing
A.~Cohan, S.~Feldman, I.~Beltagy, D.~Downey, and D.~S. Weld, ``Specter: Document-level representation learning using citation-informed transformers,'' 2020. [Online]. Available: \url{https://arxiv.org/abs/2004.07180}
\BIBentrySTDinterwordspacing

\bibitem{beltagy2020longformerlongdocumenttransformer}
\BIBentryALTinterwordspacing
I.~Beltagy, M.~E. Peters, and A.~Cohan, ``Longformer: The long-document transformer,'' 2020. [Online]. Available: \url{https://arxiv.org/abs/2004.05150}
\BIBentrySTDinterwordspacing

\bibitem{grail-etal-2021-globalizing}
\BIBentryALTinterwordspacing
Q.~Grail, J.~Perez, and E.~Gaussier, ``Globalizing {BERT}-based transformer architectures for long document summarization,'' in \emph{Proceedings of the 16th Conference of the European Chapter of the Association for Computational Linguistics: Main Volume}, P.~Merlo, J.~Tiedemann, and R.~Tsarfaty, Eds.\hskip 1em plus 0.5em minus 0.4em\relax Online: Association for Computational Linguistics, Apr. 2021, pp. 1792--1810. [Online]. Available: \url{https://aclanthology.org/2021.eacl-main.154}
\BIBentrySTDinterwordspacing

\bibitem{koike2024outfoxllmgeneratedessaydetection}
\BIBentryALTinterwordspacing
R.~Koike, M.~Kaneko, and N.~Okazaki, ``Outfox: Llm-generated essay detection through in-context learning with adversarially generated examples,'' 2024. [Online]. Available: \url{https://arxiv.org/abs/2307.11729}
\BIBentrySTDinterwordspacing

\bibitem{carlson2006rhetorical}
L.~Carlson, D.~Marcu, and M.~Okurowski, ``The rhetorical structure theory discourse treebank,'' in \emph{Proceedings of the 5th International Conference on Language Resources and Evaluation (LREC 2006)}.\hskip 1em plus 0.5em minus 0.4em\relax European Language Resources Association (ELRA), 2006.

\bibitem{prasad-etal-2008-penn}
R.~Prasad, N.~Dinesh, A.~Lee, E.~Miltsakaki, L.~Robaldo, A.~Joshi, and B.~Webber, ``The {P}enn {D}iscourse {T}ree{B}ank 2.0.'' in \emph{Proceedings of the Sixth International Conference on Language Resources and Evaluation ({LREC}'08)}, N.~Calzolari, K.~Choukri, B.~Maegaard, J.~Mariani, J.~Odijk, S.~Piperidis, and D.~Tapias, Eds.\hskip 1em plus 0.5em minus 0.4em\relax Marrakech, Morocco: European Language Resources Association (ELRA), May 2008.

\bibitem{webber2019penn}
B.~Webber, R.~Prasad, A.~Lee, and A.~Joshi, ``The penn discourse treebank 3.0 annotation manual,'' \emph{Philadelphia, University of Pennsylvania}, vol.~35, p. 108, 2019.

\bibitem{kim-etal-2020-implicit}
\BIBentryALTinterwordspacing
N.~Kim, S.~Feng, C.~Gunasekara, and L.~Lastras, ``Implicit discourse relation classification: We need to talk about evaluation,'' in \emph{Proceedings of the 58th Annual Meeting of the Association for Computational Linguistics}, D.~Jurafsky, J.~Chai, N.~Schluter, and J.~Tetreault, Eds.\hskip 1em plus 0.5em minus 0.4em\relax Online: Association for Computational Linguistics, Jul. 2020, pp. 5404--5414. [Online]. Available: \url{https://aclanthology.org/2020.acl-main.480}
\BIBentrySTDinterwordspacing

\bibitem{lin2010pdtbstyled}
Z.~Lin, H.~T. Ng, and M.-Y. Kan, ``A pdtb-styled end-to-end discourse parser,'' 2010.

\bibitem{huang2024aigeneratedtextdetectorsrobust}
\BIBentryALTinterwordspacing
G.~Huang, Y.~Zhang, Z.~Li, Y.~You, M.~Wang, and Z.~Yang, ``Are ai-generated text detectors robust to adversarial perturbations?'' 2024. [Online]. Available: \url{https://arxiv.org/abs/2406.01179}
\BIBentrySTDinterwordspacing

\bibitem{lee2024plagbenchexploringdualitylarge}
\BIBentryALTinterwordspacing
J.~Lee, T.~Agrawal, A.~Uchendu, T.~Le, J.~Chen, and D.~Lee, ``Plagbench: Exploring the duality of large language models in plagiarism generation and detection,'' 2024. [Online]. Available: \url{https://arxiv.org/abs/2406.16288}
\BIBentrySTDinterwordspacing

\bibitem{wang-etal-2024-m4}
\BIBentryALTinterwordspacing
Y.~Wang, J.~Mansurov, P.~Ivanov, J.~Su, A.~Shelmanov, A.~Tsvigun, C.~Whitehouse, O.~Mohammed~Afzal, T.~Mahmoud, T.~Sasaki, T.~Arnold, A.~Aji, N.~Habash, I.~Gurevych, and P.~Nakov, ``M4: Multi-generator, multi-domain, and multi-lingual black-box machine-generated text detection,'' in \emph{Proceedings of the 18th Conference of the European Chapter of the Association for Computational Linguistics (Volume 1: Long Papers)}, Y.~Graham and M.~Purver, Eds.\hskip 1em plus 0.5em minus 0.4em\relax St. Julian{'}s, Malta: Association for Computational Linguistics, Mar. 2024, pp. 1369--1407. [Online]. Available: \url{https://aclanthology.org/2024.eacl-long.83}
\BIBentrySTDinterwordspacing

\bibitem{fan-etal-2019-eli5}
\BIBentryALTinterwordspacing
A.~Fan, Y.~Jernite, E.~Perez, D.~Grangier, J.~Weston, and M.~Auli, ``{ELI}5: Long form question answering,'' in \emph{Proceedings of the 57th Annual Meeting of the Association for Computational Linguistics}, A.~Korhonen, D.~Traum, and L.~M{\`a}rquez, Eds.\hskip 1em plus 0.5em minus 0.4em\relax Florence, Italy: Association for Computational Linguistics, Jul. 2019, pp. 3558--3567. [Online]. Available: \url{https://aclanthology.org/P19-1346}
\BIBentrySTDinterwordspacing

\bibitem{fan2018hierarchical}
A.~Fan, M.~Lewis, and Y.~Dauphin, ``Hierarchical neural story generation,'' in \emph{Conference of the Association for Computational Linguistics (ACL)}, 2018.

\bibitem{bloomz2022}
\BIBentryALTinterwordspacing
B.~Team, ``Bloomz: A 176b parameter multilingual language model for text generation,'' \emph{arXiv preprint arXiv:2211.05180}, 2022. [Online]. Available: \url{https://arxiv.org/abs/2211.05180}
\BIBentrySTDinterwordspacing

\bibitem{sadasivan2024aigeneratedtextreliablydetected}
\BIBentryALTinterwordspacing
V.~S. Sadasivan, A.~Kumar, S.~Balasubramanian, W.~Wang, and S.~Feizi, ``Can ai-generated text be reliably detected?'' 2024. [Online]. Available: \url{https://arxiv.org/abs/2303.11156}
\BIBentrySTDinterwordspacing

\bibitem{alejandro2023multi}
\BIBentryALTinterwordspacing
A.~Alejandro and L.~Zhao, ``Multi-method qualitative text and discourse analysis: A methodological framework,'' \emph{Qualitative Inquiry}, vol.~0, no.~0, 2023. [Online]. Available: \url{https://doi.org/10.1177/10778004231184421}
\BIBentrySTDinterwordspacing

\bibitem{he2015delving}
K.~He, X.~Zhang, S.~Ren, and J.~Sun, ``Delving deep into rectifiers: Surpassing human-level performance on imagenet classification,'' in \emph{Proceedings of the IEEE International Conference on Computer Vision (ICCV)}, 2015, pp. 1026--1034.

\bibitem{agrawal2024can}
D.~Agrawal, S.~Gao, and M.~Gajek, ``Can't remember details in long documents? you need some r\&r,'' \emph{arXiv preprint arXiv:2403.05004}, 2024.

\bibitem{abedu2024llm}
S.~Abedu, A.~Abdellatif, and E.~Shihab, ``Llm-based chatbots for mining software repositories: Challenges and opportunities,'' in \emph{Proceedings of the 28th International Conference on Evaluation and Assessment in Software Engineering}, 2024, pp. 201--210.

\end{thebibliography}

\appendix

\section{Writing prompts data linguistic language features}
\label{apeendix:a}
We investigate basic linguistic features of paraWP: most commonly used words, sentence lengths, lexical diversity, token-type ratio, and further more. 

We analyzed the top 10 most frequently used terms in the HPC (e.g., like, time, back, know, said, man, see, eyes, get, around) and MGC (time, life, man, world, like, back, eyes, face, help, find in descending order of frequency) datasets. The results indicate that the most frequently used words are similar across both datasets, and their frequency of occurrence is also comparable. Overall, there is little variation in this feature.

The MGC and HPC datasets have average sentence lengths of 486.34 and 559.44, respectively, and lexical diversity (measured as the proportion of unique words to total words) of 0.498 and 0.474, indicating similar characteristics. Additionally, other linguistic features are presented in Tables \ref{tab:participle} and \ref{tab:pos-ratios-adjusted}, showing similar features and challenges for detecting them.
\begin{table}[htbp]
  \centering
  
  \begin{tabular}{lcccc}
    \hline
    Data & Avg. Sent. & ing PTCP & ed PTCP & CC \\
    \hline
    MGC & 28.04 & 13.57 & 10.84 & 16.14 \\
    HPC & 41.75 & 11.75 & 9.93 & 17.84\\
    \hline
  \end{tabular}
  \vspace{0.1cm}
  \caption{Average number of sentences, present \& past participle and connectives in the test set.}
  \label{tab:participle}
  \vspace{-0.7cm}
\end{table}

The POS tag ratios provide syntactic information, and the average lexicon diversity is calculated by dividing the number of distinct words used by the total number of words. There are no big differences in any aspect. Overall, explicit linguistic features of MGCs and HPCs are similar. It is challenging to detect MGC or HPC relying only on these.
\vspace{-0.15cm}
\begin{table}[h]
    \centering
    \small
    \begin{tabular}{lcc}
        \hline
        \textbf{POS} & \textbf{MGC Ratio} & \textbf{HPC Ratio} \\
        \hline
        Det & 0.1099 & 0.0993 \\
        Noun & 0.2424 & 0.2209 \\
        Adj & 0.0613 & 0.0674 \\
        Verb & 0.2129 & 0.2235 \\
        Adp & 0.1156 & 0.1045 \\
        Pron & 0.1217 & 0.1346 \\
        Num & 0.0051 & 0.0079 \\
        Conj & 0.0379 & 0.0349 \\
        Prt & 0.0394 & 0.0356 \\
        Adv & 0.0531 & 0.0698 \\
        X & 0.0007 & 0.0018 \\
        \hline
    \end{tabular}
    \vspace{0cm}
    \caption{POS Tag Ratios for MGC and HPC. We use universial POS tags for word classification and calculate their respective proportions. This serves as an indicator of superficial syntactic features.}
    \label{tab:pos-ratios-adjusted}
\end{table}

\end{document}